\documentclass{elsarticle}
\usepackage[english]{babel}

\usepackage[letterpaper,top=2cm,bottom=2cm,left=3cm,right=3cm,marginparwidth=1.75cm]{geometry}

\usepackage{amsmath}
\usepackage{graphicx}
\usepackage[colorlinks=true, allcolors=blue]{hyperref}

\usepackage{natbib}
\usepackage[T1]{fontenc}

\usepackage{float}
\usepackage{caption}
\usepackage{makecell}

\usepackage{multirow}
\usepackage{tabularx}
\usepackage{amssymb}
\usepackage{array}
\usepackage{booktabs}
\usepackage{subfig} 

\usepackage{afterpage}

\title{A non-invasive video-based method for individual identification of wildlife using gait dynamics}

\author[org1]{Muhammad Aamir}
\author[org2]{Matthew Wijers}
\author[org1]{Sangyun Shin}
\author[org2]{Andrew Loveridge}
\author[org1]{Andrew Markham}

\affiliation[org1]{organization={Department of Computer Science, University of Oxford},
            addressline={Wolfson Building, Parks Rd},
            city={Oxford},
            postcode={OX1 3QG},
            state={England},
            country={United Kingdom}}

\affiliation[org2]{organization={Wildlife Conservation Research Unit, Department of Biology, University of Oxford},
            addressline={Life and Mind Building, South Parks Road},
            city={Oxford},
            postcode={OX1 3EL},
            state={England},
            country={United Kingdom}}

\date{}

\begin{document}

\begin{frontmatter}

\vspace{10em}

\begin{abstract}

Gait is a distinctive behavioral characteristic that enables non-invasive individual identification without requiring physical interaction with an animal. While gait-based analysis has been extensively studied in humans, its application to wildlife remains limited due to environmental variability and the lack of scalable identification methods. This paper presents a fully automated, video-based pipeline for wildlife gait analysis and individual identification using deep spatiotemporal representation learning. The proposed pipeline uses the Segment Anything Model 3 (SAM3) to generate high-quality RGB and binary silhouette masks, robustly isolating animals from complex natural backgrounds. Segmented video sequences are processed using a convolutional neural network (ResNet18) for spatial feature extraction and a transformer-based video model (VideoPrism) for temporal motion modeling. Both models are fine-tuned using a classification objective and subsequently used as feature extractors to generate discriminative gait representations. Cosine similarity is then used to compare gait signatures, enabling similarity-based clustering of individuals without reliance on physical markings or invasive tagging. Experiments conducted on multi-source wildlife video data across multiple species demonstrate strong intra-individual consistency and clear inter-individual separation. Quantitative results using cosine similarity distributions and silhouette scores confirm the effectiveness of the proposed method. These findings demonstrate that gait dynamics provide a viable, non-invasive approach for individual identification in wildlife and highlight the potential of video-based deep learning pipelines for scalable ecological monitoring.

\end{abstract}


\end{frontmatter}

\section{Introduction}


Understanding animal locomotion provides critical insight into species behavior, biomechanics, and ecology. Although traditional gait analysis has long been applied to humans and domesticated species, its application to wildlife remains comparatively underexplored \cite{yousef2023biomechanics}. Gait analysis, broadly defined as the study of coordinated limb movement and locomotor mechanics, has long served as an essential tool in biomechanics, clinical diagnostics, and behavioral research \cite{whittle2014gait}. Historically, the foundational work by Borelli, Marey, and Muybridge established the scientific basis for understanding locomotion through mechanical modeling and chronophotography \cite{muybridge1957animals}. Advances in motion capture, inertial sensors, and image-based kinematic reconstruction have since transformed gait analysis into a quantitative discipline capable of assessing temporal, spatial, and joint-level parameters across a wide range of species \cite{kirtley2006clinical}. Although sophisticated systems are now routinely employed in laboratory settings, their application to free-ranging wildlife remains limited due to logistical, ethical, and environmental constraints.

Conventional wildlife monitoring methods, including radio collars, GPS tags, PIT tags, and physical marking, provide individual identification but often require capture, sedation, or close-range interaction, which can introduce stress and influence natural behaviors \cite{dennis2012assessing}. As ecological research increasingly emphasizes non-invasive, scalable observation techniques, video-based identification has emerged as a promising alternative \cite{norouzzadeh2018automatically}. Recent developments in computer vision and deep learning have made it feasible to extract meaningful motion signatures from unstructured video data, even under variable lighting, background complexity, and occlusion typical of natural habitats.

A critical enabling technology for such pipelines is robust foreground segmentation. Foundation models such as the Segment Anything Model 3 (SAM3) \cite{carion2025sam} offer powerful zero-shot segmentation capabilities, enabling automatic extraction of wildlife silhouettes from heterogeneous environments without species-specific training \cite{ravi2024sam}. The generation of clean RGB masks and binary silhouettes is particularly advantageous for gait analysis because it isolates the structural and temporal characteristics of the moving animal while suppressing background noise, vegetation, shadows, or camera artifacts.

Building on accurate segmentation, deep learning models have demonstrated strong performance in capturing spatiotemporal patterns that characterize individual movement styles. Convolutional architectures such as ResNet18 \cite{chen2022resnet18dnn} excel in learning spatial features related to body shape and pose, while transformer-based video models such as VideoPrism \cite{zhao2024videoprism} can encode long-range temporal dynamics and stride patterns. When fine-tuned on a classification objective using species- or individual-level data, these models learn discriminative motion representations that reflect the unique biomechanics of each animal. The resulting high-dimensional feature embeddings serve as gait signatures, suitable for downstream comparison and clustering.

Similarity based identification, particularly through metrics such as cosine similarity, provides a principled way to quantify the resemblance between gait embeddings \cite{george2020robust}. Animals exhibiting similar motion trajectories across different video sequences cluster closely in feature space, whereas distinct individuals remain well separated. This makes gait a viable form of behavioral biometrics for wildlife, independent of coat patterns, markings, or camera viewpoint constraints \cite{nixon2010human}.

The present work introduces a fully automated, non-invasive pipeline that integrates SAM3 segmentation with fine-tuned VideoPrism and ResNet18 models to extract robust gait features from wildlife videos. By using high-dimensional feature matrices and similarity based clustering, the proposed pipeline enables reliable recognition and grouping of individual animals based solely on gait dynamics. This approach offers significant potential for ecological research, conservation monitoring, and long-term behavioral studies, providing a scalable alternative to traditional identification methods in uncontrolled environments. The pipeline ultimately contributes to a broader vision of AI-driven wildlife analytics that can operate autonomously in the field.



\section{Literature Review}

\subsection{Gait Analysis and Its Evolution}

Gait analysis has evolved significantly from its early experimental origins into a sophisticated discipline that integrates biomechanics, physiology, and computational modeling. The foundational work of Borelli, Marey, and Muybridge \cite{borelli2015borelli,braun2012eadweard} demonstrated that systematic observation of locomotion provides valuable insights into coordination, efficiency, and musculoskeletal function. Throughout the 20th century, advances in motion capture, cine film, and high speed photography enabled the quantification of joint trajectories, ground reaction forces, and temporal gait parameters in both humans and animals. Contemporary gait analysis using a combination of imaging technologies, pressure platforms, wearable inertial sensors, and three dimensional kinematic reconstruction to characterize locomotion under laboratory and natural conditions \cite{ahmadi20163d}. These tools allow for a detailed assessment of stride patterns, limb coordination, and pathological deviations. Although most applications historically focus on human health and rehabilitation, recent efforts are increasingly explore gait in domestic and exotic animals, contributing to veterinary medicine, biomechanics, and behavioral ecology \cite{hutchinson2021evolutionary}. Despite technological progress, applying gait analysis to wildlife remains challenging due to uncontrolled environments, making automated vision-based systems an attractive emerging alternative.

\subsection{Feature Extraction and Similarity-Based Identification}

Effective gait-based identification depends on extracting discriminative spatiotemporal features that capture the subtle dynamics of an individual specie locomotion \cite{liu2018learning}. Traditional methods relied on handcrafted features such as gait energy images (GEI), contour trajectories, or joint-angle time series \cite{han2005individual}. Although useful, these representations often struggle with viewpoint variation and environmental noise. Deep learning has transformed feature extraction by enabling models to learn robust identity cues directly from video sequences. Convolutional neural networks (CNNs), recurrent architectures, and more recently transformer-based models encode both appearance and motion, resulting in high-dimensional embeddings that generalize well across contexts. Once extracted, these embeddings can be compared using similarity metrics, with cosine similarity being particularly effective at measuring the angular alignment between feature vectors. This allows sequences belonging to the same individual to cluster closely in feature space while maintaining separation from others. Such similarity-driven pipelines have been widely successful in face recognition, speaker verification, and re-identification tasks, and are increasingly employed in animal identification research, where consistent labeling is difficult and traditional biometric cues may be unavailable.


\subsection{Gait as a Biometric Feature}

Gait has become a powerful biometric trait due to its unobtrusive, behavior-based nature and its inherent distinctiveness between individuals \cite{purish2023gait,cattin2002biometric}. Unlike facial or fingerprint biometrics, gait can be captured at a distance without cooperation, making it valuable in surveillance, security, and identity verification. The uniqueness of gait arises from the interplay of skeletal structure, neuromuscular patterns, learned habits, and anatomical proportions, all of which shape characteristic stride rhythms and postural dynamics. Studies have demonstrated that factors such as stride length, cadence, body orientation, and joint angle trajectories form stable identity signatures under normal conditions. However, gait is also susceptible to covariates, including speed, clothing, footwear, fatigue, and emotional or cognitive states, which can alter appearance-based features or temporal patterns. These challenges have motivated the development of robust machine learning and model based representations capable of capturing invariant identity cues \cite{parashar2022robust}. As deep learning evolved, gait biometrics expanded beyond handcrafted silhouettes to high-dimensional spatiotemporal embeddings, enabling significant improvements in recognition accuracy under unconstrained conditions.



\subsection{Deep Learning for Animal Gait Recognition}

Although gait recognition research is dominated by human studies, interest in extending these techniques to animals has grown rapidly, driven by applications in veterinary diagnostics, livestock management, and wildlife ecology \cite{kuhl2013animal,boyd2005biometric}. Early efforts relied on marker-based motion capture or manual annotation to study animal locomotion, limiting scalability and applicability in natural settings. With the rise of deep learning and pose estimation frameworks such as DeepLabCut \cite{mathis2018deeplabcut}, OpenPose \cite{martinez2019openpose}, and YOLO-based detectors \cite{raza2025analyzing}, researchers can now automatically estimate keypoints and extract motion patterns from videos of animals ranging from rodents to large mammals. These methods have enabled automated lameness detection in cattle, gait abnormality assessment in dogs, and species-level gait classification. However, wildlife poses unique challenges including irregular movement patterns, variable lighting, occlusion, complex backgrounds, and scarcity of labeled datasets. To overcome these constraints, recent work has explored silhouette-based feature extraction, spatiotemporal convolutional networks \cite{zeng2023video}, and transformer-based video models that can learn directly from raw or segmented video data \cite{arnab2021vivit}. Such approaches show a strong potential for individual identification and long-term monitoring of free-ranging animals. Table \ref{tab:gait_literature} presents a brief literature of the existing studies on gait analysis using deep machine learning.

\begin{table}[ht]
\centering
\caption{Existing approaches used in gait analysis across humans and animals}
\label{tab:gait_literature}
\begin{tabular}{llp{2.5cm}p{2.5cm}p{2.5cm}p{3.0cm}}
\toprule
\textbf{Article} & \textbf{Subject} & \textbf{Dataset} & \textbf{Method} & \textbf{Purpose} & \textbf{Taxonomy} \\

\midrule

\cite{nixon2010human} &
Human & SOTON, CMU MoBo & GEI, model-based features &Person identification & Biometric (GEI) \\

\cite{chao2019gaitset} & Human & CASIA-B & Set-based deep model & View-invariant gait recognition & Deep Learning (Silhouette) \\

\cite{cui2023multi} & Human & CASIA-B, OU-MVLP & Transformer-based fusion network & Multi-view gait recognition & Transformer based Gait Models \\

\cite{he2016deep} & General & ImageNet & ResNet (CNN backbone) &
Spatial feature extraction (used in gait methods) & CNN Backbone \\

\cite{makihara2021gait} & Human & CASIA database &
GEI + PCA/LDA & Baseline gait biometrics & Biometric (GEI) \\

\cite{sheth2023gait} & Human & OU-ISIR & Silhouette CNN & Gait verification &
Biometric (CNN) \\

\cite{zhang2019graph} & Human & CASIA-B, KS20 & Skeleton GCN & Pose-based gait recognition & Graph/Skeleton \\

\cite{cao2019openpose} & Human & COCO, MPII & 2D pose estimation &
Joint trajectory extraction & Pose Estimation \\

\cite{martinez2017simple} & Human & Human3.6M & 3D pose regression network &
Joint kinematics for gait-style analysis & Pose-based Methods \\

\cite{mathis2018deeplabcut} & Animals & Various lab datasets & Deep pose estimation & Kinematics of animal locomotion & Animal Pose Estimation \\

\cite{shrestha2017gait} & Animals & Equine gait lab data &
Marker-based kinematics & Veterinary diagnosis & Medical / Marker-based \\

\cite{gill5262411machine} & Animals & Custom canine dataset & CNN + motion features & Breed identification via gait & Animal-specific CNN \\

\bottomrule
\end{tabular}
\end{table}

\section{The Proposed Pipeline}

The proposed pipeline is presented in Figure \ref{fig:pipeline} and the overall pipeline consists of the following primary stages:

\begin{figure}[h]
\includegraphics[width=\linewidth]{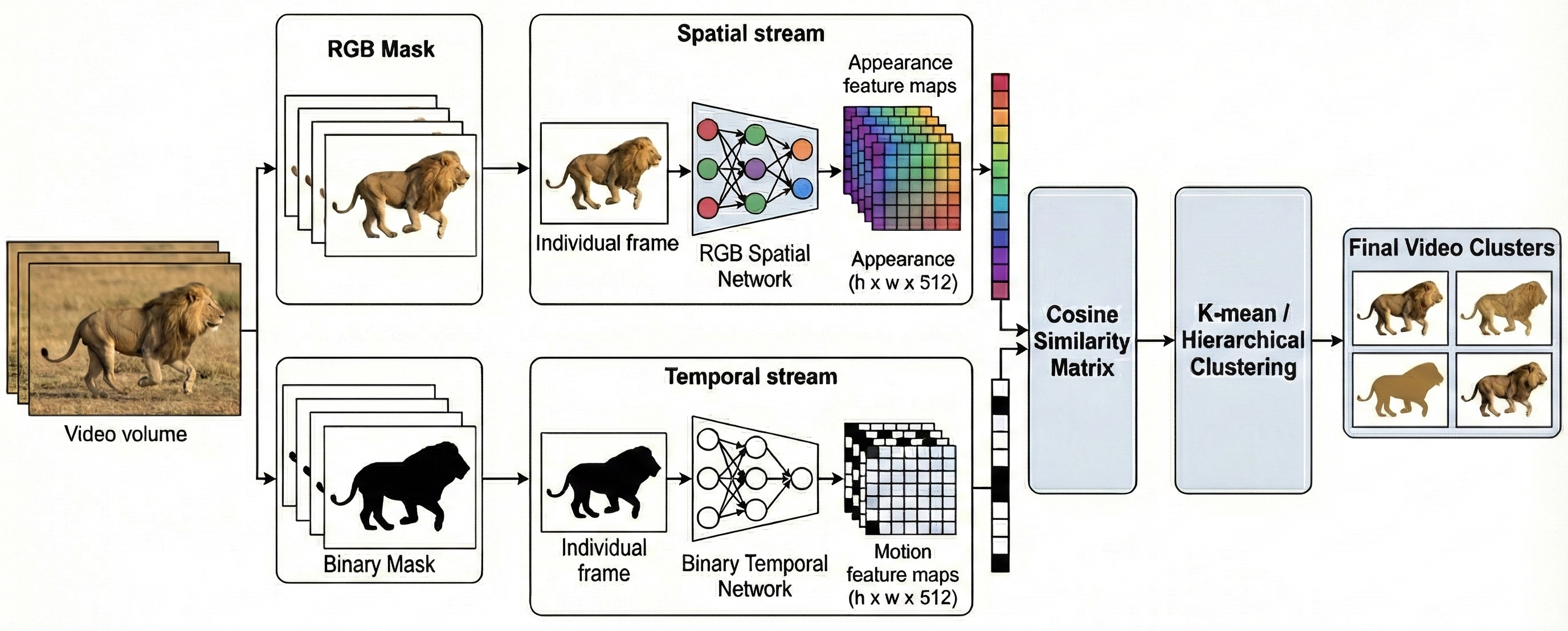}
\caption{The proposed video based pipeline for unique identification using gaits analysis}
\label{fig:pipeline}
\end{figure}

\subsection{Data Acquisition and Preprocessing}

For this study, a diverse and representative dataset of wildlife gait videos was assembled from multiple publicly accessible and privately collected sources. The goal was to capture natural walking behaviors across different species, environments, and camera conditions to ensure robust model training and feature extraction. Publicly available platforms such as YouTube, the NESTLER Wild Animal channel, and free nature documentary footage were used to collect videos of animals walking in open habitats, semi-captive reserves, and mixed-terrain environments. These sources provided natural diversity in lighting, background complexity, and animal pose orientation.

To supplement this material with high quality field recordings, additional data was collected at Longleat Safari Park (UK). This custom dataset enabled controlled observation of animal locomotion from multiple angles and distances while maintaining non-invasive conditions. In total, 185 video sequences were curated, representing 5 species: Camel, Lion, Giraffe, Zebra, and Hyena. Videos of a single species contain 4 to 6 individual animals and each individual has 5 to 20 video snippets. It should be noted that some of the videos of the same individual are captured from different sides. Each species contributes distinct morphological and gait characteristics, ranging from the rolling gait of camels to the diagonal walking patterns of giraffes, ensuring that the dataset captures a wide spectrum of locomotive dynamics.
All videos were standardized to consistent frame rates and resolutions. Segments containing clear walking behavior were extracted and trimmed. This multi-source dataset provides a robust foundation for segmentation, feature learning, and gait based individual identification.

\subsection{Segmentation}

Accurate foreground segmentation is essential for isolating the animal’s locomotion patterns from complex natural environments. To achieve this, each video frame is processed using the SAM3, which generates both a binary silhouette mask and an appearance-preserving RGB mask. SAM3 improves temporal consistency through internal motion-aware refinement, ensuring that contours remain stable across consecutive frames even in environments with dynamic movement, variable illumination, or partial occlusions. The resulting masks capture the structural characteristics and body movement essential for gait analysis. These segmented sequences are then passed to the feature extraction models. The following mathematical formulation describes the segmentation and mask generation process

Let the input video be a sequence of RGB frames:
\[
\mathcal{I} = \{I_t\}_{t=1}^{T}, \qquad I_t : \Omega \rightarrow \mathbb{R}^3 .
\]

SAM3 produces a binary silhouette mask:
\[
B_t = S(I_t; \theta_S, p_t),
\]
where
\[
B_t(x) =
\begin{cases}
1, & \text{if pixel } x \text{ belongs to the animal},\\[4pt]
0, & \text{otherwise}.
\end{cases}
\]

The RGB segmentation mask is:
\[
R_t(x) = B_t(x)\, I_t(x), \qquad x \in \Omega .
\]

Temporal consistency is enforced by minimizing:
\[
\{\hat B_t\} = \arg\min_{\{\tilde B_t\}}
\left(
\sum_{t=1}^{T} \mathcal{E}_{\text{data}}(\tilde B_t, I_t)
+ \lambda_{\text{temp}} \sum_{t=2}^{T}
\mathcal{E}_{\text{temp}}(\tilde B_{t-1}, \tilde B_{t})
\right).
\]

\[
\mathcal{E}_{\text{data}}(\tilde B_t, I_t)
= \sum_{x \in \Omega}
\ell_{\text{seg}}\big(\tilde B_t(x), B_t(x)\big),
\]
where $\ell_{\text{seg}}$ is a pixel-wise segmentation loss.

\[
\mathcal{E}_{\text{temp}}(\tilde B_{t-1}, \tilde B_t)
=
\sum_{x \in \Omega}
\left|
\tilde B_t(x) -
\tilde B_{t-1}\big(x + w_{t-1 \to t}(x)\big)
\right|,
\]
where $w_{t-1 \to t}$ is the optical flow field.

\subsection{Deep Feature Extraction}

After segmentation, the masked sequences are processed using two complementary deep learning models: ResNet18 \cite{chen2022resnet18dnn} for spatial features and VideoPrism \cite{zhao2024videoprism} for temporal motion cues. ResNet18 captures appearance-independent structural information such as contour shape and limb configuration, while VideoPrism captures sequential gait dynamics including stride periodicity and joint coordination. The combination of these models provides a rich spatiotemporal representation of each animal’s gait signature. Both models were fine-tuned using a classification objective to make their embeddings more discriminative. The fused embeddings serve as the basis for downstream similarity computation. The mathematical formulation below describes how spatial and temporal features are extracted and merged.

 $s_t$ encodes the shape/appearance at time t
\[
s_t = f_R(R_t; \theta_R)
\]

$v_t$ summarizes gait motion around frame t
\[
v_t = f_V(R_{t-\tau+1:t}; \theta_V)
\]

Combined spatiotemporal feature vector
\[
h_t = \operatorname{concat}(s_t, v_t)
\]

$\bar h$ represents the gait signature of the entire video
\[
\bar h = \frac{1}{T} \sum_{t=1}^{T} h_t
\]

Fine-tuning objective: cross-entropy classification loss
\[
\mathcal{L}_{cls}
= -\sum_i \log
\frac{\exp(W\bar h_i + b)_{y_i}}
{\sum_c \exp(W\bar h_i + b)_c}
\]

\subsection{Feature Matrix and Similarity Computation}
After obtaining the spatiotemporal feature embeddings, each video sequence was converted into a structured feature matrix, where rows represent temporal segments (frames or frame windows) and columns represent feature dimensions derived from deep learning models. These matrices capture both static anatomical cues and dynamic gait patterns.
To quantify the similarity between gait signatures of different animals, we computed the cosine similarity between the embeddings. Cosine similarity evaluates the angular distance between high-dimensional vectors, making it robust to variations in feature magnitude and particularly suitable for comparing deep representations. A similarity matrix was generated for all pairwise combinations of videos, revealing the degree of similarity between each individual's gait patterns. 

This method allows the pipeline to operate independently of visual markings, background context, or environmental variations. The mathematical formulation below describes how the feature matrix is structured and how similarity is computed.

Feature matrix for the $i_{th}$ video where rows represent frames, columns are feature dimensions
\[
H^{(i)} =
\begin{bmatrix}
h^{(i)}_1 \\
h^{(i)}_2 \\
\vdots \\
h^{(i)}_{T_i}
\end{bmatrix}
\]

Sequence-level embedding
\[
\bar h^{(i)} = \frac{1}{T_i}\sum_{t} h^{(i)}_t
\]

Cosine similarity between video i and j
\[
\mathrm{sim}(i,j)
=
\frac{
\langle \bar h^{(i)}, \bar h^{(j)} \rangle
}{
\|\bar h^{(i)}\|_2\,
\|\bar h^{(j)}\|_2
}
\]

Construct affinity matrix between all videos
\[
A_{ij} = \mathrm{sim}(i,j)
\]

\subsection{Clustering and Identification}

The final stage of the pipeline involves clustering individual animals based on the similarity of their gait signatures. Using the cosine similarity scores computed earlier, an unsupervised learning approach was used to group video sequences without manual labels. Techniques such as hierarchical clustering and k-means clustering were tested to evaluate different grouping behaviors and cluster granularity. Hierarchical clustering provides intuitive dendrogram structures that reveal inter-animal relationships and relative similarity distances, while k-means offers clearer, centroid-based cluster assignments. In both approaches, sequences belonging to the same individual consistently formed tight clusters, demonstrating the discriminative power of the learned gait embeddings. The mathematical expressions below describe the clustering process based on the affinity matrix computed earlier.


\[
D_{ii} = \sum_{j=1}^{N} A_{ij}.
\]

\[
L_{\mathrm{sym}} = I - D^{-1/2} A D^{-1/2}.
\]

Compute the $k$ smallest eigenvectors and run k-means on rows of U to obtain cluster labels.
\[
U \in \mathbb{R}^{N \times k}.
\]

Hierarchical clustering distance:
\[
d_{ij} = 1 - A_{ij}.
\]

Silhouette coefficient:
\[
a(i) = \frac{1}{|C_i|-1}
\sum_{\substack{j \in C_i \\ j \ne i}}
d_{ij},
\]
\[
b(i) = \min_{C \ne C_i}
\frac{1}{|C|}
\sum_{j \in C}
d_{ij},
\]

\[
s(i) = 
\frac{
b(i) - a(i)
}{
\max\{a(i), b(i)\}
}.
\]

Identification decision rule:
\[
\hat y =
\arg\max_{g}
\operatorname{sim}\big(\bar h^{(p)}, \bar h^{(g)}\big),
\]

Or match if

\[
\max_g \operatorname{sim} \ge \tau.
\]

\section{Results and Discussion}

The proposed gait analysis pipeline was evaluated on a collection of wildlife videos featuring multiple species and individual animals observed under varying environmental conditions. The primary goal was to assess whether the fine-tuned VideoPrism and ResNet18 models could produce discriminative gait features that reliably cluster instances of the same animal while maintaining separability between different individuals. Across all evaluated species, the results consistently indicate that gait-based spatiotemporal embeddings capture identity-specific locomotion cues, even in the presence of background variability and moderate viewpoint changes.


\subsection{Qualitative Observations}

\begin{figure}[h!]
\includegraphics[width=\linewidth]{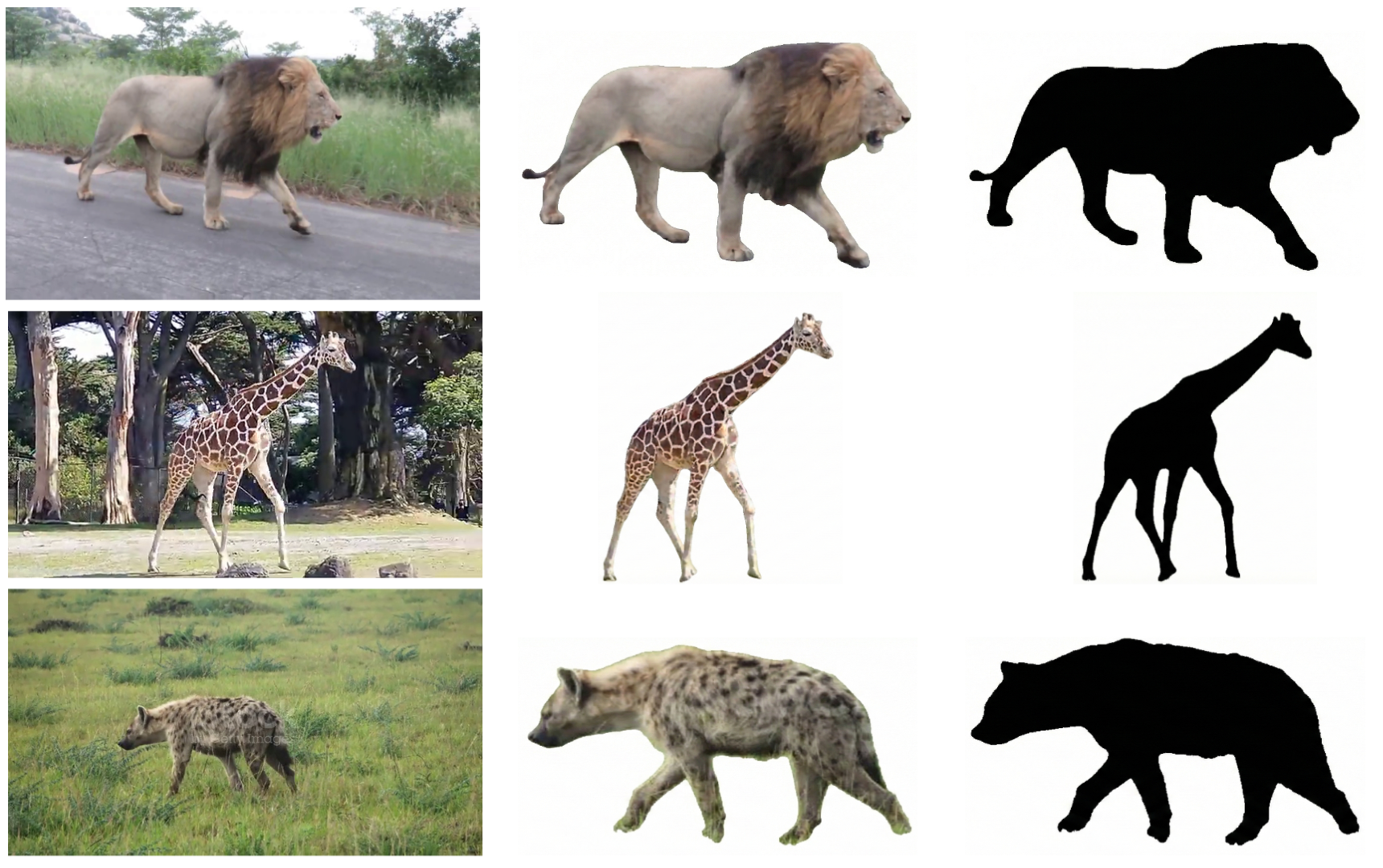}
\caption{Masks generation using SAM3 from random frames of Lion, Giraffe, and Hyena videos. First column represent the original frame, second column is the RGB masks of animal and the third column represents the binary mask.}
\label{fig:masks}
\end{figure}

Visual inspection of the segmented outputs from SAM3 confirmed that the model effectively extracted clean and consistent RGB and binary masks, even under moderate occlusion and background complexity. For animals with irregular body shapes or non-uniform fur patterns, the dual-mask strategy helped maintain shape continuity, which was critical for subsequent gait representation learning. Figure \ref{fig:masks} shows random examples of mask extraction using SAM3, highlighting the robustness of segmentation across species with diverse morphologies.

\begin{figure}[h!]
    \centering
    
    \subfloat[Camel]{\label{fig:kmeans_sub4}
        \includegraphics[width=0.33\textwidth]{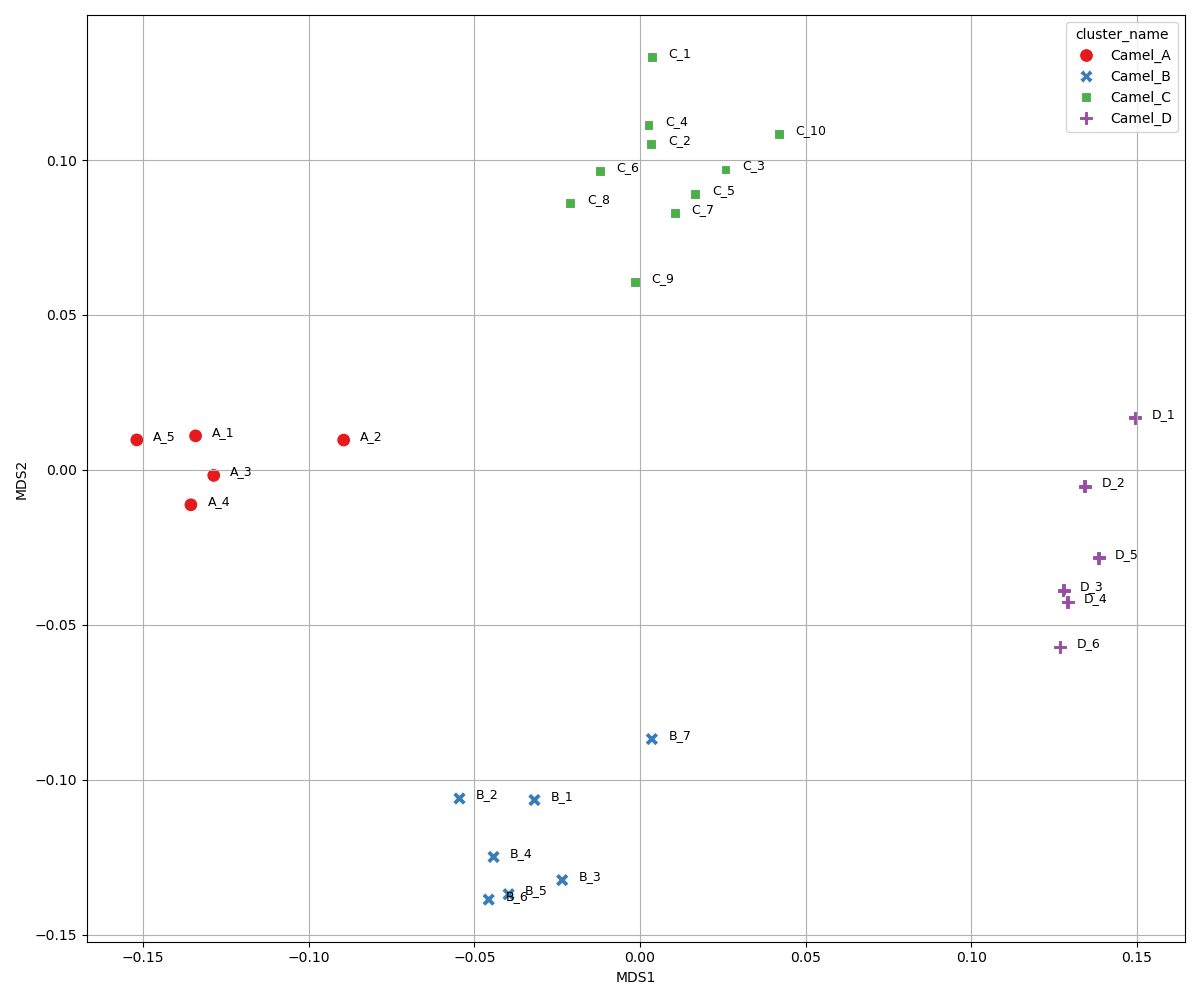}}
    \subfloat[Giraffe]{\label{fig:kmeans_sub2}
        \includegraphics[width=0.33\textwidth]{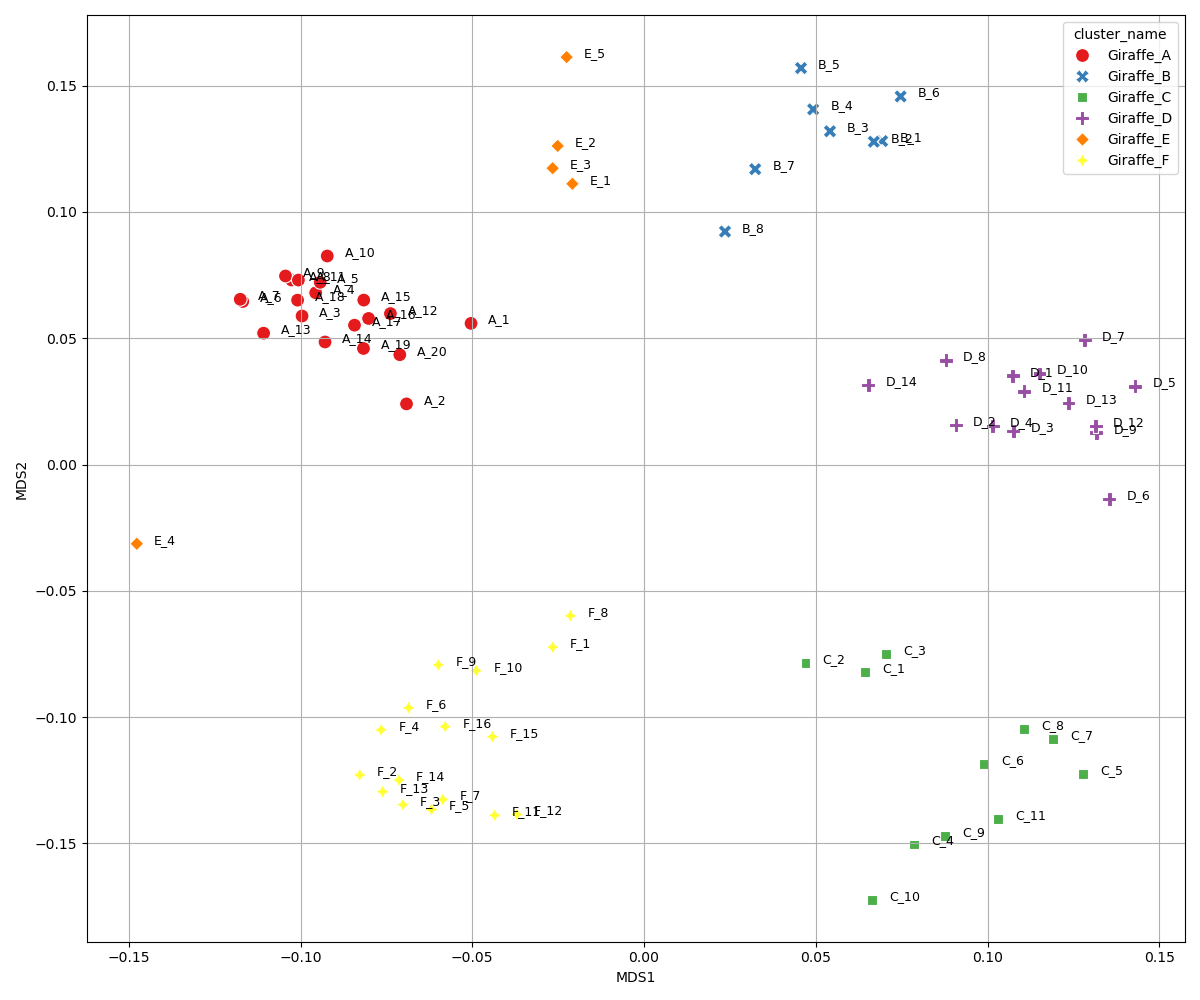}}
    \subfloat[Lion]{\label{fig:kmeans_sub1}
        \includegraphics[width=0.33\textwidth]{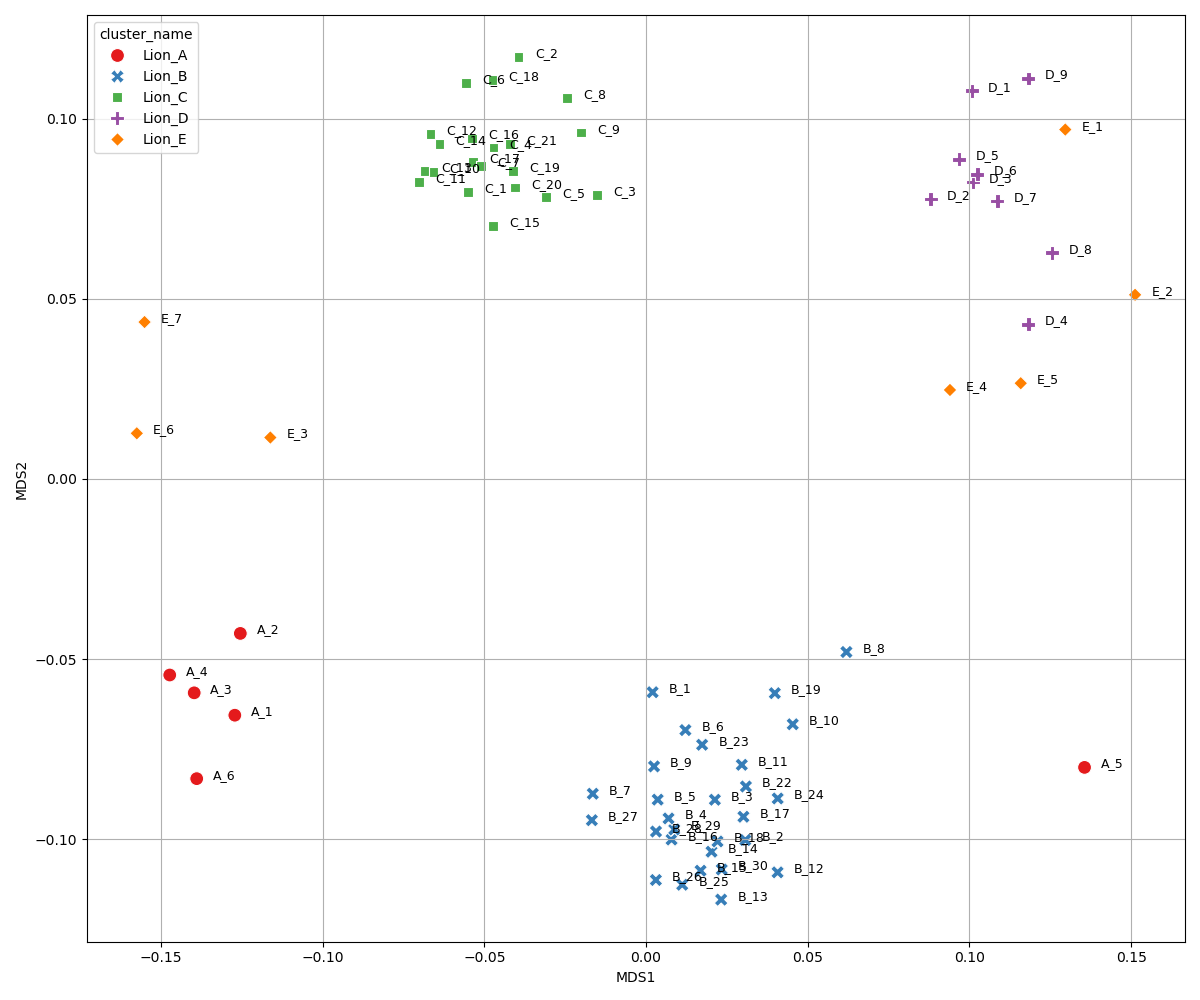}}
    
    \vspace{-0.5em}
    \subfloat[Zebra]{\label{fig:kmeans_sub5}
        \includegraphics[width=0.33\textwidth]{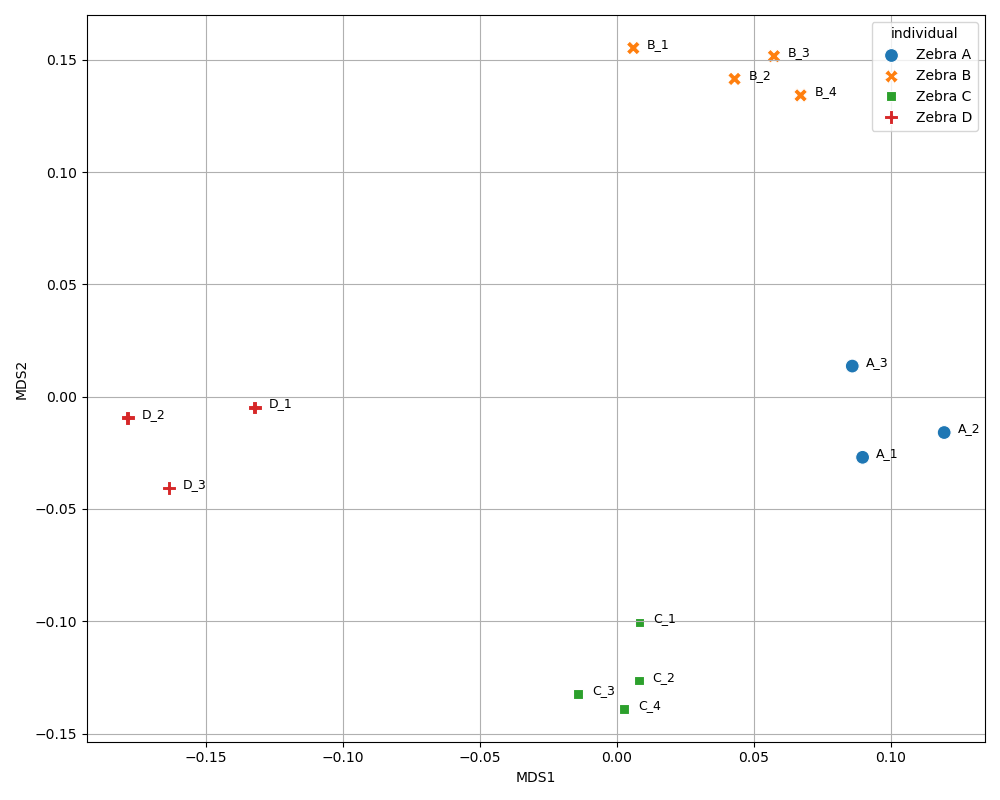}}
    \subfloat[Hyena]{\label{fig:kmeans_sub3}
        \includegraphics[width=0.33\textwidth]{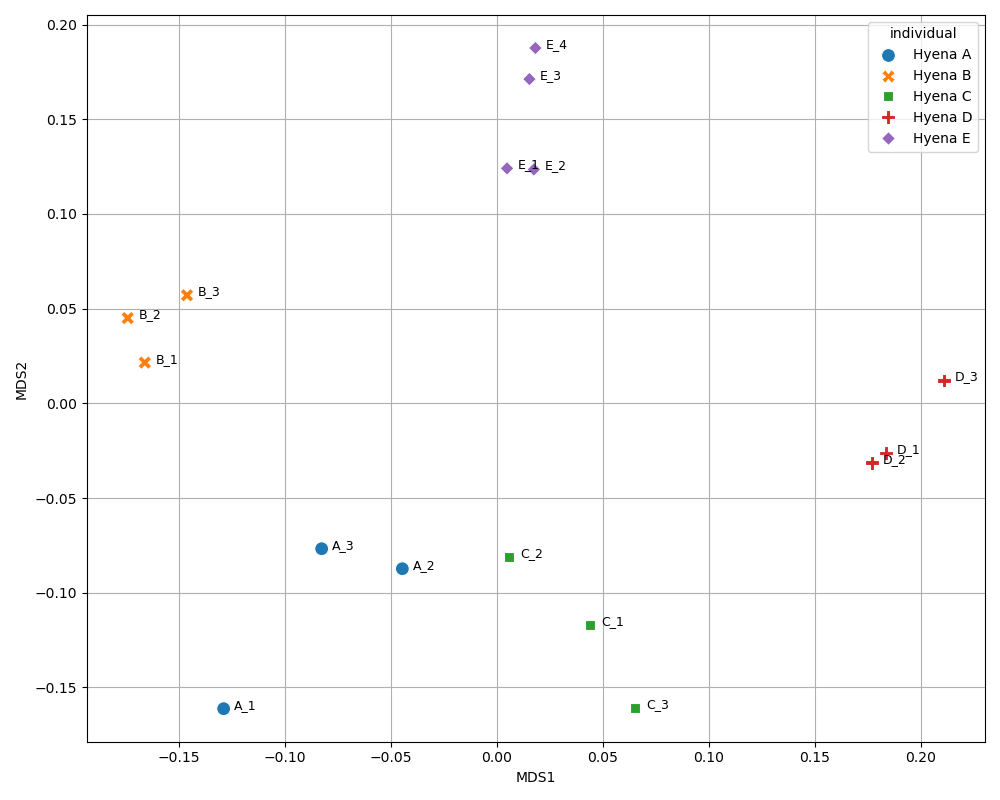}}
    
    


    \caption{K-mean clustering using Multidimensional Scaling (MDS) to project the cosine similarity into 2D space}
    \label{fig:cluster}
\end{figure}

The gait embeddings extracted by the fine-tuned models were projected into two-dimensional space using t-SNE and UMAP visualizations. These projections show distinct clusters corresponding to individual animals. Video sequences from the same individual, captured at different times, consistently formed compact clusters, whereas different animals were well separated. Figure \ref{fig:cluster} illustrates the output of k-means clustering at the individual level after projecting the cosine similarity matrix into 2D space using Multidimensional Scaling (MDS). This visualization demonstrates that the learned representations encode identity-specific locomotion patterns rather than superficial appearance cues.

\begin{figure}[h!]
    \centering
    
    \subfloat[Camel]{\label{fig:dendrogram_sub4}
        \includegraphics[width=0.45\textwidth]{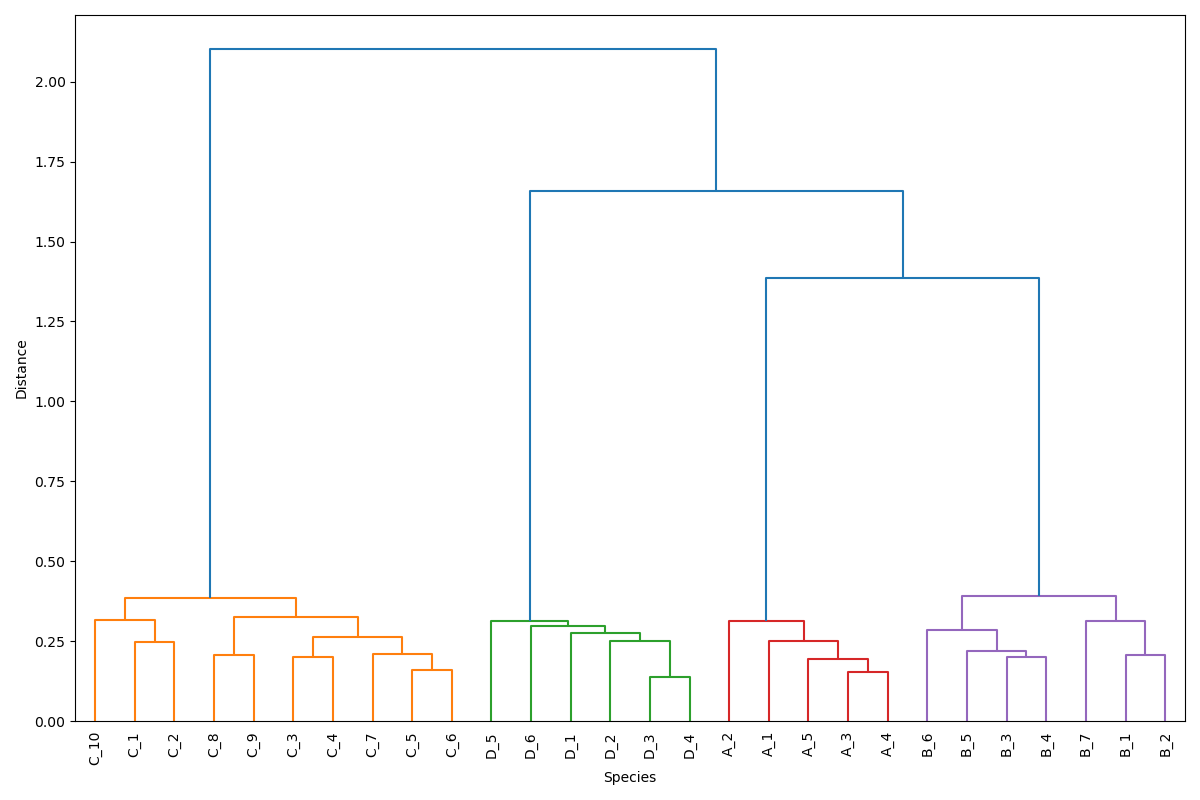}}
    \subfloat[Giraffe]{\label{fig:dendrogram_sub2}
        \includegraphics[width=0.45\textwidth]{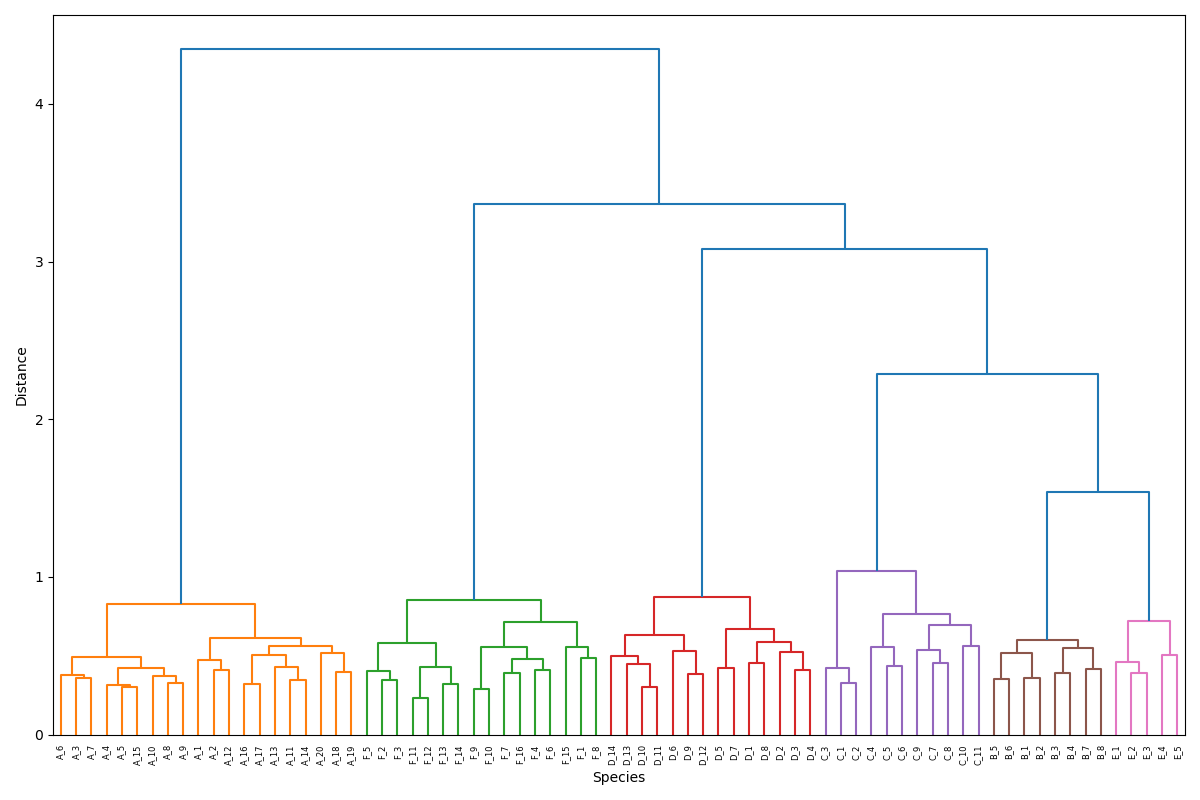}}
        
\vspace{-0.5em}
    
    \subfloat[Zebra]{\label{fig:dendrogram_sub5}
        \includegraphics[width=0.45\textwidth]{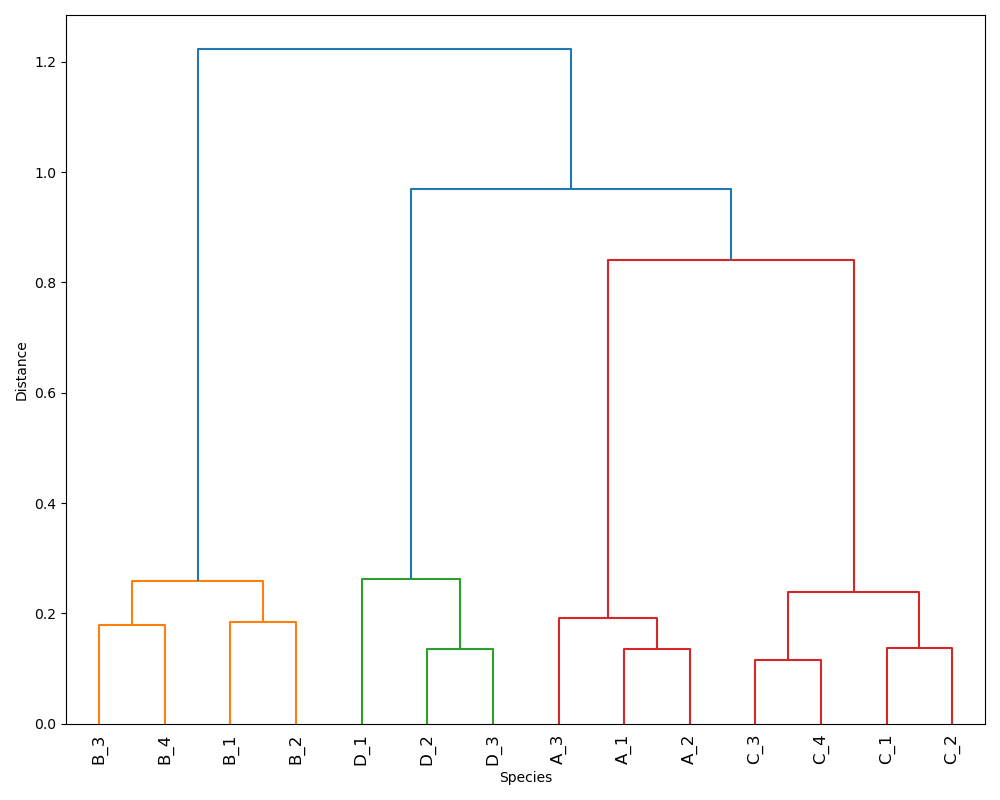}}
    \subfloat[Hyena]{\label{fig:dendrogram_sub3}
        \includegraphics[width=0.45\textwidth]{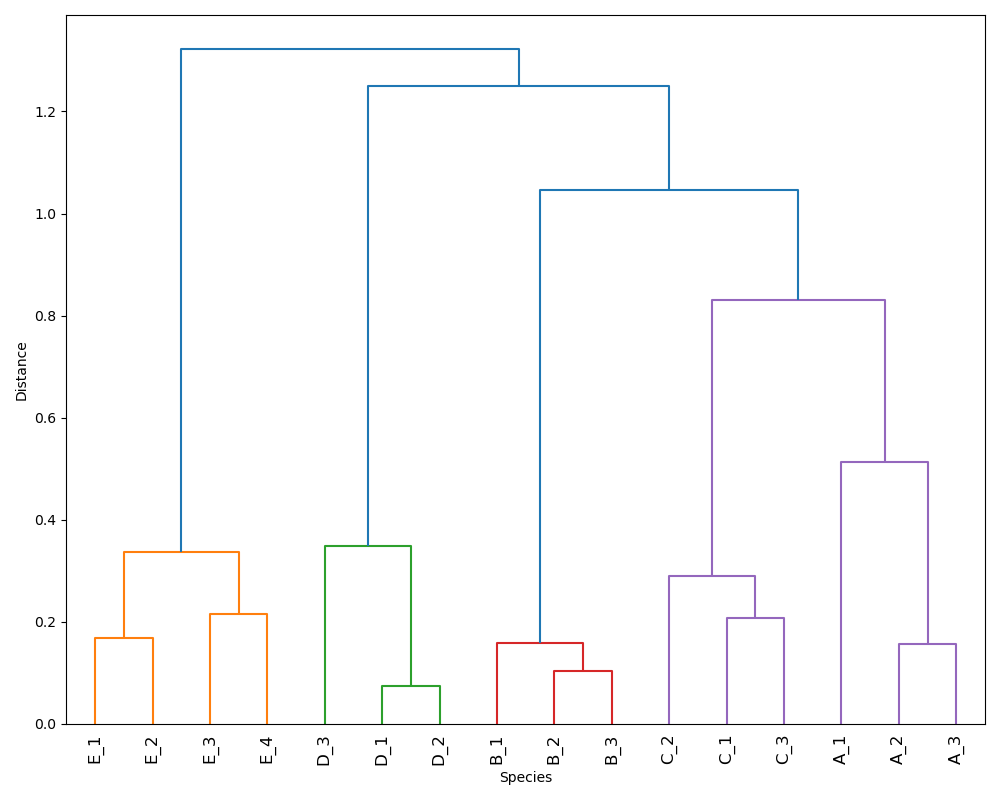}}
\vspace{-0.5em}

    \subfloat[Lion]{\label{fig:dendrogram_sub1}
        \includegraphics[width=0.5\textwidth]{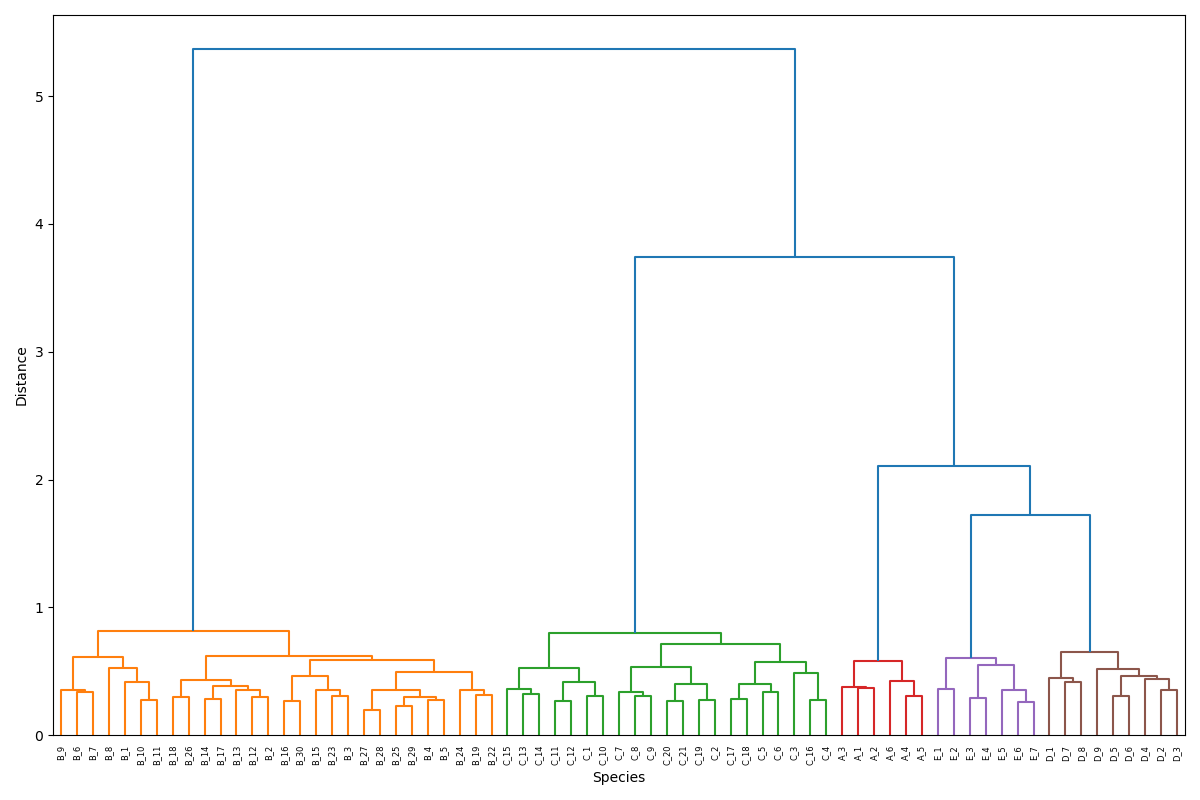}}
        
    \caption{Hierarchical clustering dendrograms illustrating gait-based similarity between individual animals for each species. Shorter linkage distances indicate higher gait similarity, while larger separations reflect distinct individual gait signatures.}
    \label{fig:dendograms}
\end{figure}

In addition to the k-means results, Figure \ref{fig:dendograms} presents hierarchical clustering dendrograms derived from the cosine similarity of the gait embeddings for Zebra, Giraffe, Camel, Lion and Hyena. The dendrogram structures reveal clear hierarchical separation between individuals, with shorter linkage distances observed within the same individual and larger distances between different individuals. This hierarchical organization further supports the stability and discriminative nature of the learned gait signatures.

\vspace{-0.5em}

\subsection{Quantitative Evaluation}

To quantitatively assess the effectiveness of the proposed gait-based identification pipeline, cosine similarity matrices were computed for all pairwise combinations of video sequences within each species. Across all evaluated species, the similarity matrices exhibited a pronounced block-diagonal structure, indicating strong intra-individual consistency and clear separation between different individuals. Intra-animal cosine similarity values ranged between $0.91$ and $0.99$, while the majority of inter-animal similarities remained below $0.80$. This separation suggests that the learned spatiotemporal embeddings reliably capture identity-specific gait characteristics rather than incidental visual features.

Cluster quality was further evaluated using the silhouette coefficient, which measures both compactness and separability of clusters. An average silhouette score of $0.78$ was obtained across species, confirming that the resulting clusters are well defined. Species with more regular and periodic gait patterns, such as zebras and giraffes, tended to achieve higher silhouette scores, whereas species exhibiting more variable locomotion dynamics, such as hyenas, showed slightly reduced but still robust performance.

Figure~\ref{fig:similarity} presents confusion matrices derived from cosine similarity scores for each species, providing a fine-grained view of individual-level performance. While most individuals demonstrate strong intra-class similarity, certain individuals show reduced similarity due to viewpoint variation. For example, ``Camel D'' in Figure~\ref{fig:similarity_sub3} exhibits lower intra-similarity because recordings were captured from opposing lateral viewpoints. Similar trends are observed for ``Lion E'' and ``Giraffe E'' in Figures~\ref{fig:similarity_sub4} and \ref{fig:similarity_sub2}, respectively. These results highlight the sensitivity of gait embeddings to viewpoint changes and underscore the importance of consistent viewing geometry.

\afterpage{%
\begin{figure}[p]
    \centering

    \subfloat[Camel]{\label{fig:similarity_sub3}
        \includegraphics[width=0.5\textwidth]{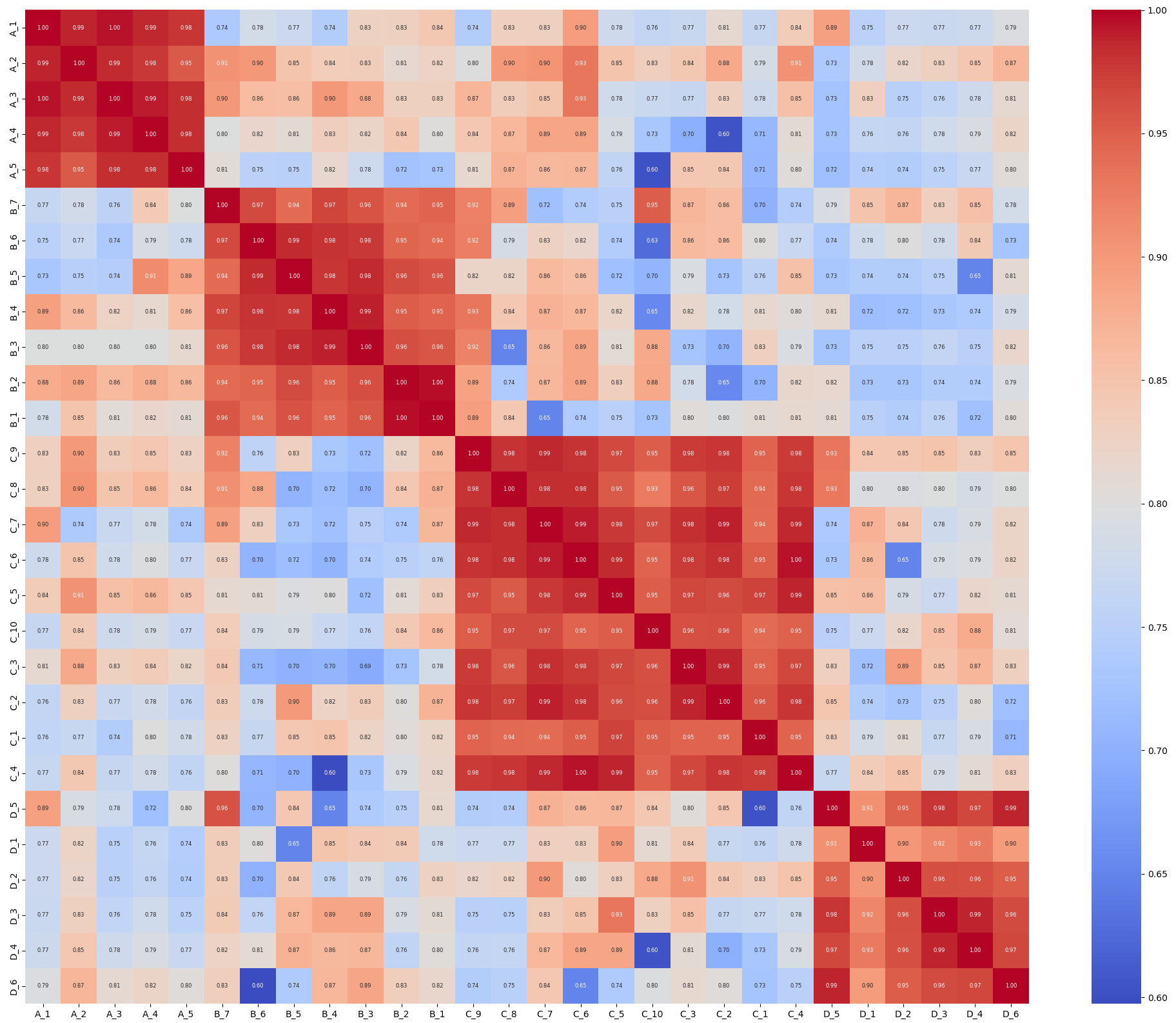}}
    \subfloat[Giraffe]{\label{fig:similarity_sub2}
        \includegraphics[width=0.5\textwidth]{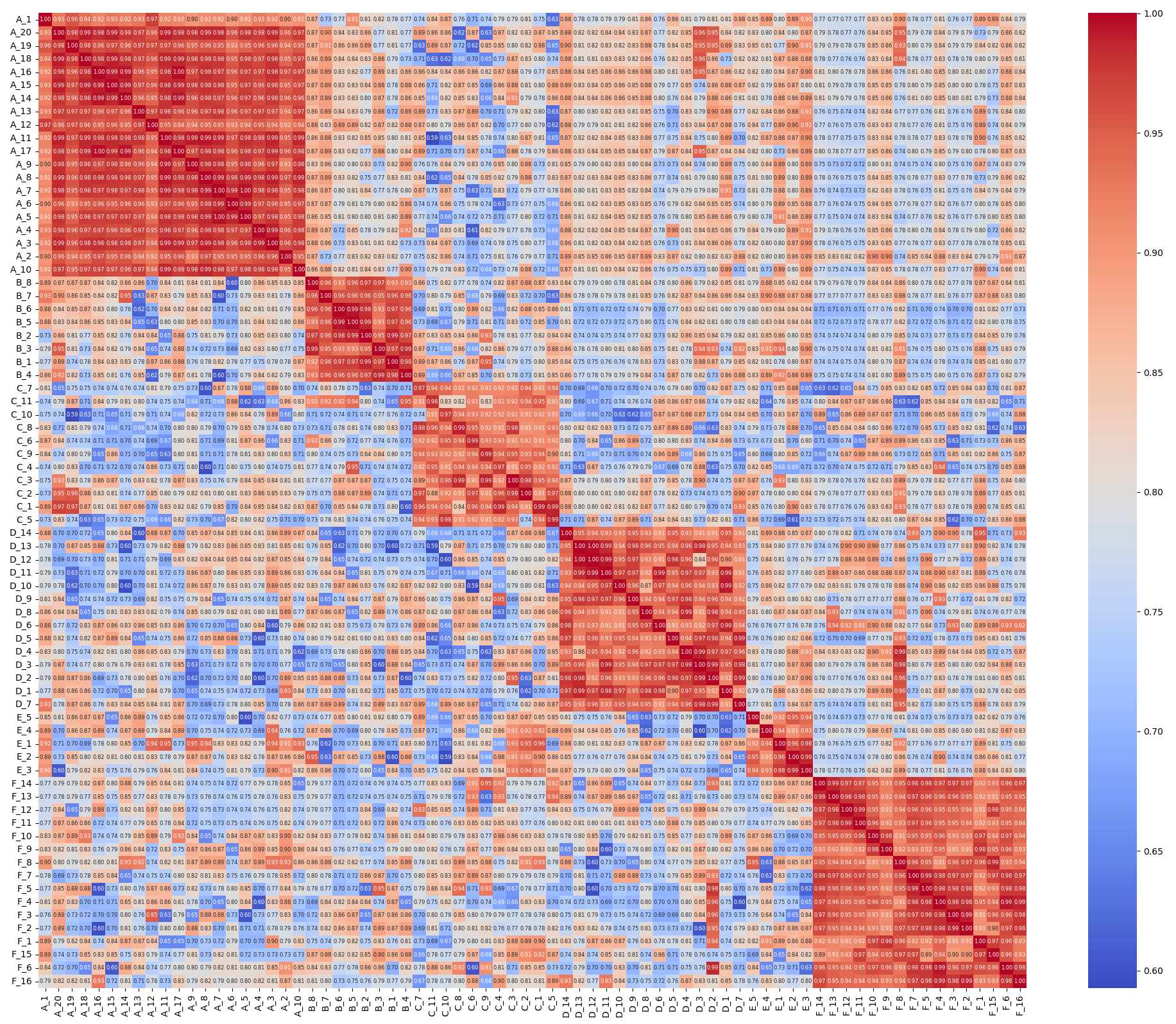}}
    
    \vspace{-0.5em}
    
    \subfloat[Zebra]{\label{fig:similarity_sub1}
        \includegraphics[width=0.5\textwidth]{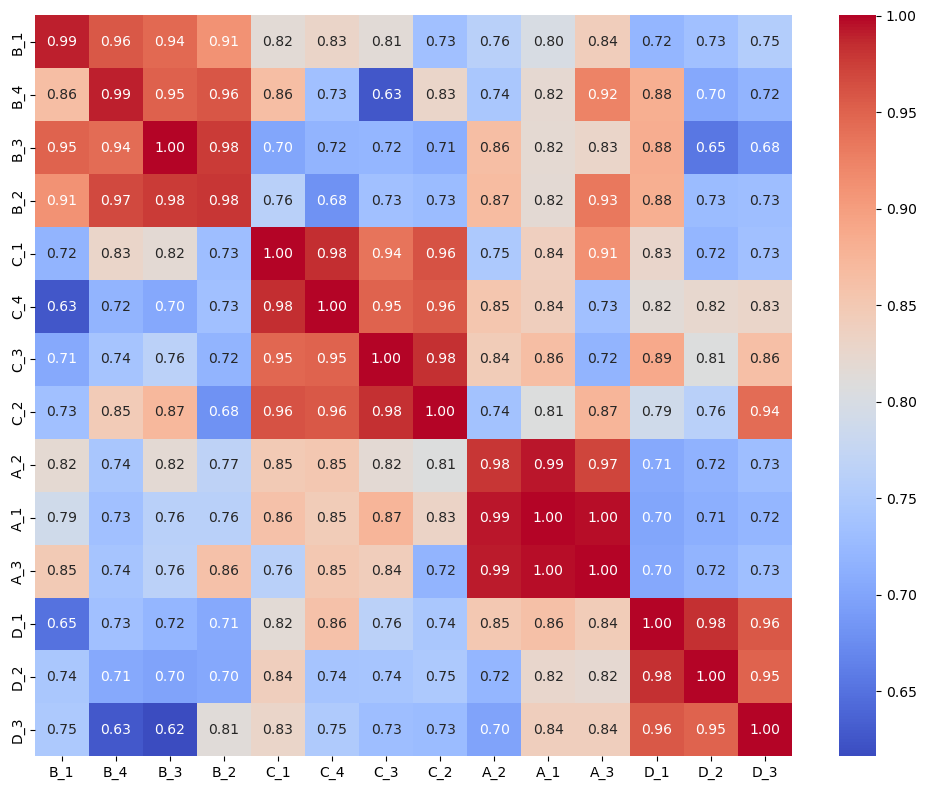}}
    \subfloat[Hyena]{\label{fig:similarity_sub5}
        \includegraphics[width=0.5\textwidth]{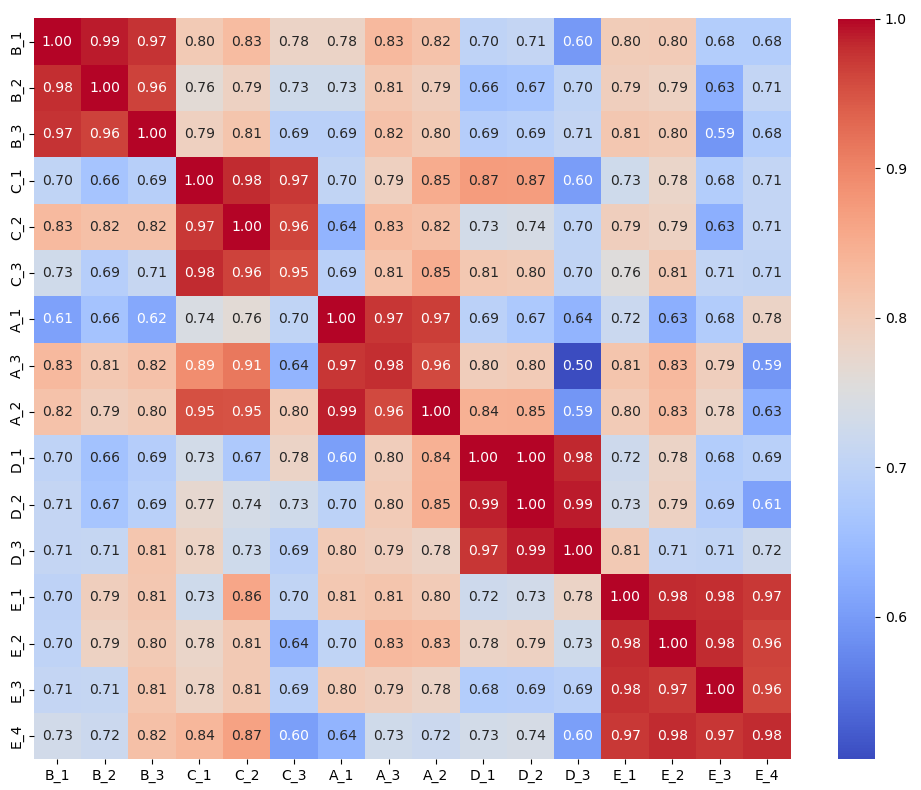}}
    
    \vspace{-0.5em}
    
    \subfloat[Lion]{\label{fig:similarity_sub4}
        \includegraphics[width=0.6\textwidth]{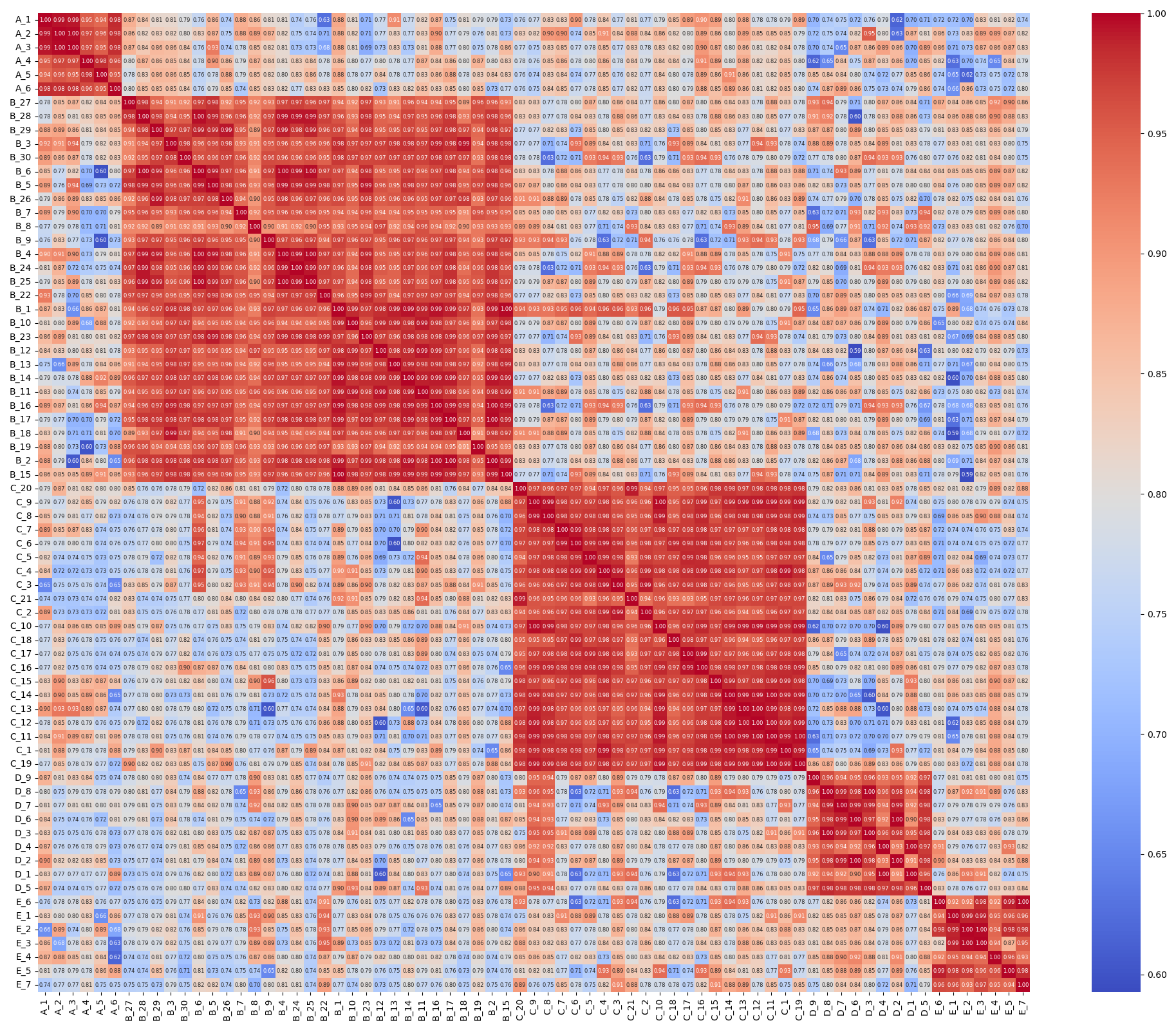}}
        
    \caption{Similarity in the form of confusion matrix for the all five species}
    \label{fig:similarity}
\end{figure}
\clearpage
}

\subsection{Comparative Analysis}

To analyze the contribution of individual components in the proposed pipeline, ablation experiments were conducted using spatial-only and temporal-only feature representations. When using ResNet18 alone, which captures spatial and structural cues such as body shape and silhouette geometry, the model achieved moderate clustering performance with an average intra-animal cosine similarity of $0.78$. Although spatial features provide useful identity information, they are insufficient to fully capture the dynamic aspects of gait, leading to increased overlap between individuals with similar morphology.

Using VideoPrism alone, which focuses on temporal motion dynamics, improved intra-sequence consistency to an average similarity of $0.82$. This indicates that temporal features such as stride rhythm and limb coordination play a critical role in identity discrimination. However, reliance on temporal features alone occasionally resulted in confusion between individuals exhibiting similar gait rhythms, particularly within the same species.

The fusion of spatial features from ResNet18 and temporal features from VideoPrism produced the strongest performance, with intra-animal cosine similarity values ranging from $0.91$ to $0.98$ across species. This improvement demonstrates that spatial structure and temporal dynamics provide complementary information for gait-based identification. Quantitative evidence for this behavior is summarized in Table~\ref{tab:comparative_similarity}, where all species exhibit consistently high mean intra-animal similarity values (above $0.95$), while maintaining substantially lower inter-animal similarity.

Notably, hyenas achieved the highest intra-animal similarity ($0.976$), suggesting highly consistent gait patterns across recordings, whereas giraffes showed slightly increased inter-animal similarity ($0.795$), likely due to their elongated limb structure and reduced lateral stride variability. Camels and lions exhibited higher inter-animal similarity values ($0.800$ and $0.808$, respectively), which aligns with observed viewpoint variations and similar body proportions across individuals.

Overall, these results reinforce the importance of spatiotemporal feature fusion and demonstrate that the proposed pipeline generalizes effectively across species with diverse morphologies and locomotion styles.

\begin{table}[ht]
\centering
\caption{Mean intra-animal and inter-animal cosine similarity values across species.}
\label{tab:comparative_similarity}
\begin{tabular}{lcc}
\toprule
\textbf{Species} & 
\textbf{ Intra-Animal} & 
\textbf{ Inter-Animal} \\
\midrule
Zebra    & 0.961 & 0.777 \\
Hyena   & 0.976 & 0.747 \\
Camel   & 0.966 & 0.800 \\
Lion    & 0.966 & 0.808 \\
Giraffe & 0.954 & 0.795 \\
\bottomrule
\end{tabular}
\end{table}

\section{Conclusion and Implications}

This research work introduced a biometric framework for gait based wildlife identification using segmentation-driven deep spatiotemporal learning from video data. By integrating segmentation using SAM3 with spatial feature extraction using ResNet18 and temporal modeling via VideoPrism, this pipeline effectively captures Identifiable gait signatures from unconstrained wildlife videos. The use of cosine similarity and unsupervised clustering enabled reliable differentiation between individuals without dependence on physical appearance cues, manual labeling, or invasive identification methods. Experimental results across multiple species demonstrated high intra-individual similarity and strong inter-individual separability, validating the discriminative capability of the learned gait embeddings.

From a biometric perspective, this study extends gait recognition beyond human subjects and demonstrates its applicability to wildlife in real-world conditions. Importantly, all videos used in this work describe animals walking laterally across the camera field of view (left-to-right or right-to-left). This controlled viewpoint ensures consistent observation of steps kinematic, limb coordination, and body movements, which are critical for reliable gait-based identification. Under these conditions, the proposed approach shows strong potential for long-term monitoring of individual animals in conservation areas, safari parks, and semi-wild environments.

The implications of this research are significant for wildlife conservation and behavioral ecology. The proposed framework enables scalable, non-invasive biometric identification using readily available video data, reducing the need for tagging, collars, or close human–animal interaction. Moreover, the pipeline is species-agnostic and benefits from foundation models that generalize well across diverse morphologies and environments, making it suitable for deployment in heterogeneous ecological settings.

In the future, we plan to extend the framework to handle more challenging viewpoints, including frontal and rear-facing walking sequences, where gait cues are less pronounced. Incorporating multi-view learning, 3D gait reconstruction, and self-supervised or contrastive learning strategies may further improve robustness and generalization. Expanding the dataset to include varying terrains, walking speeds, and behavioral states will also be essential to advance gait-based wildlife biometrics toward real-world deployment.




\bibliographystyle{elsarticle-num-names}
\bibliography{refs}

\end{document}